\documentclass[fleqn,10pt]{wlscirep}
\usepackage[utf8]{inputenc}
\usepackage[T1]{fontenc}
\usepackage{lineno}
\usepackage{comment}
\usepackage{graphicx}
\usepackage{subcaption}

\usepackage{soul}

\title{Real-time, low-cost multi-person 3D pose estimation}

\author[1]{Alice~Ruget}
\author[1]{Max~Tyler}
\author[2]{Germán~Mora-Martín}
\author[1]{Stirling~Scholes}
\author[1]{Feng~Zhu}
\author[2]{Istvan~Gyongy}
\author[3]{Brent~Hearn}
\author[1]{Steve~McLaughlin}
\author[1]{Abderrahim~Halimi}
\author[1,*]{Jonathan~Leach}
\affil[1]{School of Engineering and Physical Sciences, Heriot-Watt University, Edinburgh, EH14 4AS, UK}
\affil[2]{School of Engineering, Institute for Integrated Micro and Nano Systems, The University of Edinburgh, Edinburgh, EH9 3FF, UK}
\affil[3]{Imaging Sub-group, STMicroelectronics, Edinburgh, EH3 5DA, UK}
\affil[*]{Jonathan Leach (j.leach@hw.ac.uk)}

\begin{abstract}

The process of tracking human anatomy in computer vision is referred to pose estimation, and it is used in fields ranging from gaming to surveillance. Three-dimensional pose estimation traditionally requires advanced equipment, such as multiple linked intensity cameras or high-resolution time-of-flight cameras to produce depth images. However, there are applications, e.g.~consumer electronics, where significant constraints are placed on the size, power consumption, weight and cost of the usable technology. Here, we demonstrate that computational imaging methods can achieve accurate pose estimation and overcome the apparent limitations of time-of-flight sensors designed for much simpler tasks. The sensor we use is already widely integrated in consumer-grade mobile devices, and despite its low spatial resolution, only 4$\times$4 pixels, our proposed Pixels2Pose system transforms its data into accurate depth maps and 3D pose data of multiple people up to a distance of 3 m from the sensor. We are able to generate depth maps at a resolution of 32$\times$32 and 3D localization of a body parts with an error of only $\approx$10 cm at a frame rate of 7 fps.  This work opens up promising real-life applications in scenarios that were previously restricted by the advanced hardware requirements and cost of time-of-flight technology.

\end{abstract}
\begin{document}

\flushbottom
\maketitle
\thispagestyle{empty}

\twocolumn

\section*{Introduction}

Pose estimation is the process of locating the position of human body parts via analysis of images, videos, and sensor data. Accurate tracking of human anatomy is important in several areas, including activity recognition in gaming \cite{:Zanfir}, gesture identification in consumer electronics \cite{Farooq:18}, behavioural analysis in medical monitoring \cite{Dama:16,Mathis:18}, as well as form and functional analysis in professional sports \cite{Moeslund:06}.  Three-dimensional pose estimation from depth images or depth videos has been performed across many different domains:  fall detection of elderly \cite{Xiong_20, Bian_15,Serpa_20}, medical diagnosis \cite{Qingqiang_21}, assistance in physical therapy  \cite{Physical_therapy_19, fallrecoveryassist16}, monitoring of patient sleep \cite{monitoringsleep_18}, sport coaching \cite{Park_17}, interaction with robots \cite{Lewandowski_19}, and general action recognition \cite{Yang_19,Keçeli_17,Liu:2017}. As the application areas for pose estimation span a wide range, so too does the technology used for it.  For example, the most accurate pose estimation uses markers or multiple sensors that are tracked in three dimensions. Accurate 3D tracking can also be obtained using high-resolution depth images or triangulation from multiple linked intensity cameras.

While advanced technology is known to provide accurate pose estimation, it is also desirable to have accurate tracking from the simplest possible technology.  Approaching the problem from this perspective opens up opportunities where cost, size, and weight are significant considerations, e.g. the consumer electronics market, autonomous and self-driving vehicles, and airborne vehicles such as drones.  Here we show that a simple, small, and cost-effective time-of-flight sensor with only 4 $\times$ 4 pixels contains sufficient data for 3D tracking of multiple human targets.  

Very accurate pose estimation can be achieved by placing markers on the body. For example, inertial markers that record motion by combining data from different sensors such as accelerometers, gyroscopes, or magnetometers can recover accurate body poses \cite{Baldi-20,Yun_06} and can be used in combination with images \cite{Marcard_18,Gilbert_19}. They have been developed, for example, for clinical applications \cite{Bodysensor_sports} and for tracking posture during sport \cite{Bodysensors4,BodySensors3}. Marker-based pose estimation gives the most accurate results, but these technologies are expensive, time-consuming to use, and the requirement to wear sensors means that they are not practical for general applications.  Accurate poses can also be estimated using several linked cameras viewing a scene from different angles \cite{Vlasic_08, Carranza_03, Iskakov_19,Mehrizi_18}. Reflective markers placed on the body or face are also commonly used for animation and special effects in computer games or films \cite{Body_markers_avatar}. These approaches are very reliable, but it is desirable to have methods that do not require multiple cameras or use any markers.

Three-dimensional pose estimation from single point-of-view intensity images is an attractive alternative to labeled tracking because such images are easy to obtain \cite{Poppe_07,Moeslund_06}.  However, 3D pose estimation in this manner is extremely challenging due to depth ambiguities and occlusions from objects. Recent algorithms based on machine learning networks have achieved 3D pose estimation from single RGB images, demonstrating the reconstruction of multiple people that is robust to occlusions, in real-time, and in both controlled and uncontrolled environments \cite{Bala_20, Kidzinski_20, Rogez_20, Mehta_20, Benzine_21,Liu_20,Agarwal_06,Chen_17}. Ref.~\cite{Wang2020} contains a comprehensive  collection of resources on pose estimation from RGB images.

An alternative to using RGB images is using depth images to reconstruct 3D poses \cite{Zhou_20, Zhang_20, Moon_18}. Depth images provide a considerable advantage since they already contain 3D information, however, more advanced sensors are required to record depth.  For fast depth data acquisition, two main technologies are used: time-of-flight (ToF) cameras or structured light sensors. ToF cameras use a pulse of light to illuminate a target and a detector records the returned light. Structured light sensors project a pattern of light onto the scene and the depth measurement is based on triangulation \cite{survey_depthimagery_13}. Recent research has developed techniques to retrieve high-resolution depth images from a single-pixel depth detector, therefore opening new perspectives in terms of cost and speed. Structured illumination has been used to reconstruct high-resolution depth images from a single-pixel detector at high frame rates \cite{Sun_16} and from indirect light measurements of static scenes \cite{Zhang_15}. However, the hardware requirements for structured illumination make it impossible at this stage to be integrated into high-scale marketed consumer devices. 

An attractive solution to the requirement to use structured illumination for single-pixel depth imaging was proposed by Turpin {\it et al.}~who demonstrated that the rich temporal data from a single point in space contains information that could be converted to depth images via reconstruction with a neural network \cite{Turpin:20}. This work shows an important proof-of-concept that good spatial resolution can be retrieved from temporal data rather than from the detector's spatial structure. Estimating depth from a single pixel appeared to be a heavily ill-posed inverse problem, yet the authors show that the use of a static background overcomes this apparent constraint.

\begin{figure*}[ht!]
\centering
\includegraphics[width=\textwidth]{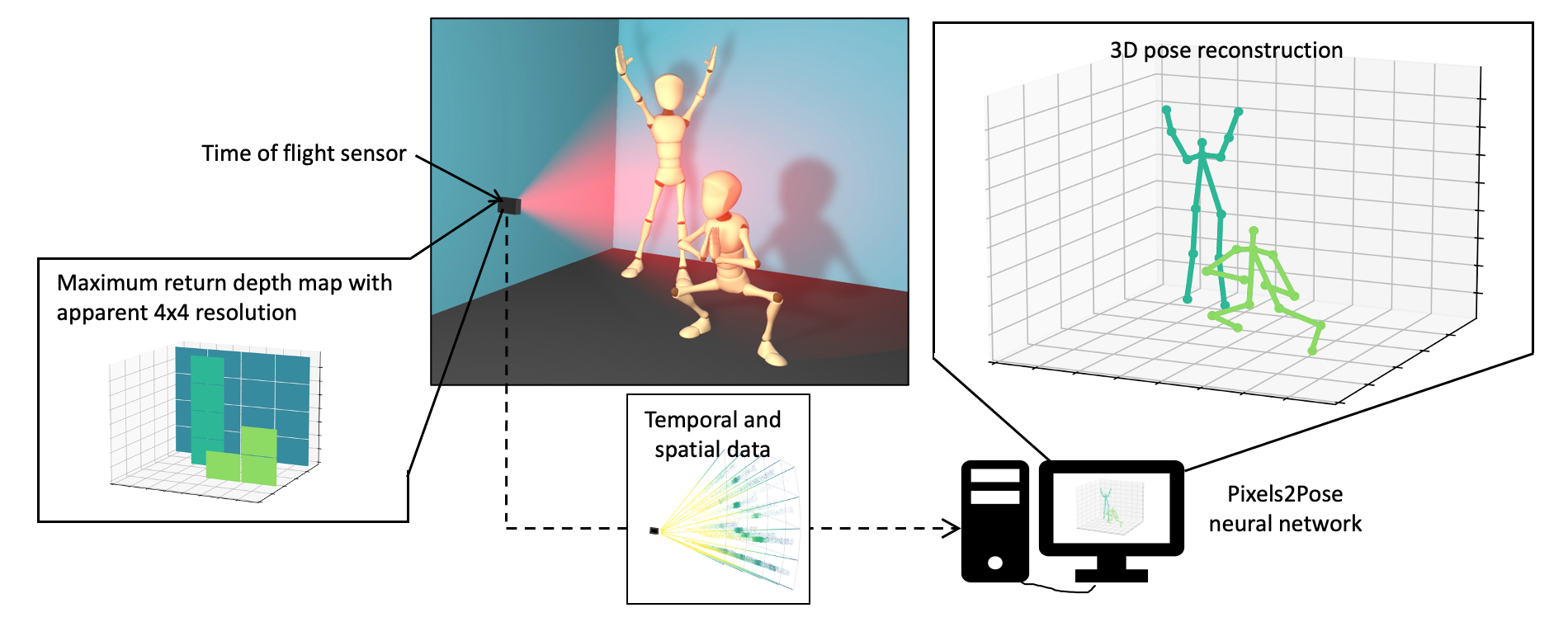}
\caption{\label{fig:intro}
\textbf{Schematic of the Pixels2Pose system.}  A small, cost-effective time-of-flight sensor illuminates a scene and generates histogram data with a spatial resolution of 4$\times$4 (x, y).  This data is passed to the Pixels2Pose network to generate accurate pose reconstruction in 3D. }
\end{figure*}

Computational imaging methods are known to provide powerful tools to extract and convert information between different modalities, provided that the input signal is rich and the task is sufficiently restricted.  The work in reference \cite{Turpin:20} was developed further to show depth imaging of people using multi-path temporal echoes from radar, sonar, and lidar data \cite{Turpin_21}, and the poses of humans that are behind walls can be estimated using data obtained at radar frequencies \cite{Zhao_18, Zhao_18_2D}. Additionally, networks that use multiple input data source have been used for data fusion to increase the resolution of depth images originating from the temporal histogram data from single-photon detector array sensors and intensity images \cite{Wetzstein_20, Lindell:2018, Sun_20,Ruget:21}.

On the other hand, small, cost-effective ToF depth detectors with very few pixels have been developed for commercial purposes and are designed for applications such as auto-focus assist or obstacle detection in smartphones and drones.  While such sensors only have a few pixels, they have rich temporal information, and Callenberg {\it et al.}~recently demonstrated a range of applications that are significantly enhanced by use of the full ToF histogram data from a cheap commercial SPAD sensor \cite{Callenberg2021CheapSPAD}. This work highlights the increasing range of applications that can be delivered from such a ToF sensor.

Our work builds upon the core ideas of image enhancement using neural networks and processing the full histogram data obtained from cost-effective ToF SPAD sensors. The use of the full ToF histogram data from few pixels is key to the success of this work, and we show that generating depth images from a cheap, simple depth sensor can be achieved at high frame rates. Not only can we reconstruct depth images, but these images also have sufficient resolution to perform accurate 3D pose estimation of multiple targets.  Crucially, as the sensor has multiple pixels, our system solves a more constrained ill-posed problem and therefore the pre-trained network works in a range of different environments.

\section*{Results}


\begin{figure*}
\centering
\includegraphics[width=\textwidth]{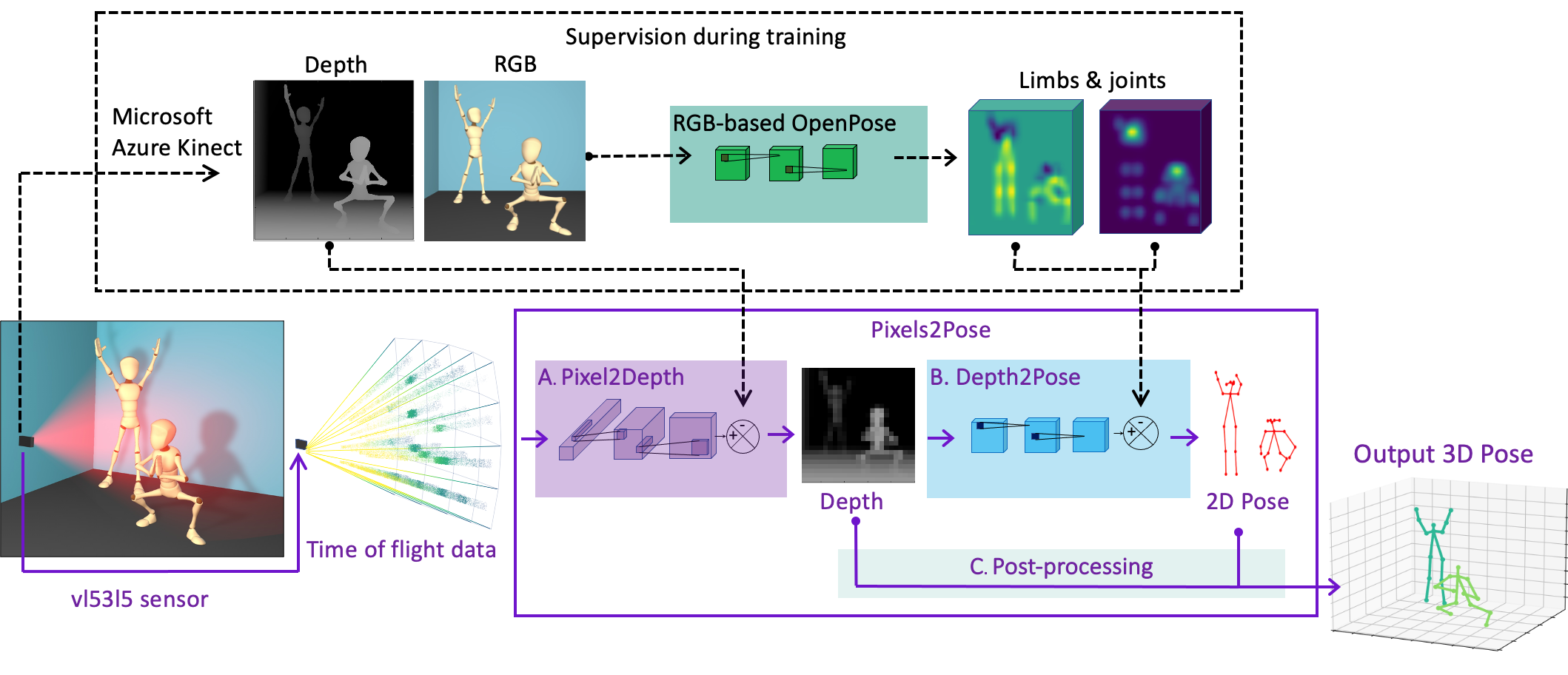}
\caption{\label{fig:postprocessing}
\textbf{Overview of Pixels2Pose along with the supervision used for training.} The bottom part displays the Pixels2Pose system. The ToF data of the sensor is passed through a series of three steps to reconstruct the 3D Pose: A. the network Pixels2Depth returns a high resolution (HR) depth map from the histogram data; B. the network Depth2Pose processes the HR depth map to return 2D poses; C. the HR depth map and the 2D poses are combined to produce 3D poses. 
The top part displays the system used for the training of the networks Pixels2Depth and Depth2Pose. A Microsoft Azure Kinect DK camera is used to provide the labels corresponding to the sensor data. For Pixels2Depth, the high-resolution depth images of the Kinect are used as labels. For Depth2Pose, the RGB image is processed through OpenPose \cite{OpenPose} to get the 2D pose labels.}
\end{figure*}

\subsection*{Overview of the system}
The Pixels2Pose system utilizes a small sensor to illuminate a scene and generate ToF histogram data of size $4\times4\times144$ (x,y,t). This data is then passed to a neural network that has been trained to recover the poses of multiple people in three dimensions.  The training stage of Pixels2Pose uses high-resolution depth and intensity images obtained from a Microsoft Kinect sensor and the RGB-based pose network OpenPose \cite{OpenPose}.  Despite the apparent low spatial resolution, after the supervised training, our proposed Pixels2Pose system transforms the sensor’s rich ToF data into accurate 3D pose data of multiple people. A schematic of the system is shown in Figure \ref{fig:intro}.

Our Pixels2Pose system is made of two neural networks: one that estimates depth from measured histograms and one inspired from the network OpenPose \cite{OpenPose} that creates 2D poses using heatmaps of joints and part affinity fields. Our final step consists of superimposing the two outputs to render a 3D pose. We demonstrate continuous real-time video at a frame rate of 7 fps. Our approach can be adopted widely in a range of systems due to the simplicity of the underlying technology.

\subsection*{Sensor}
The key sensor for our work is the vl53l5 single-photon avalanche diode (SPAD) sensor manufactured by STMicroelectronics. The sensor illuminates the scene with 940 nm light pulses, and its SPAD detectors record the time of arrival of photons reflected as histograms of photon counts. The field of view is 60 degrees diagonal, the maximum range is 3 meters and the frame rate is 10fps. The dimensions of the sensor are 4.9mm$\times$2.5mm$\times$1.6mm, the spatial resolution is only 4$\times$4 pixels, and the temporal resolution is 144 time-bins, each separated by 125 ps. The data is cropped to 100 time-bins so that there are no unwanted artefacts from objects in the background.  We can establish the main depth in each pixel, i.e., a single depth associated with the time-bin showing maximum return of photons.  This provides a  4$\times$4 maximum return depth map.
A visual representation of the temporal and spatial data from the vl53l5 sensor and its corresponding maximum return depth map are shown in Fig. \ref{fig:intro}.

\subsection*{Pixels2Pose Network}
The proposed Pixels2Pose system takes the raw data of the sensor as its input, i.e. the 4$\times$4 histograms of 100 time-bins each, generates a higher resolution depth map, and then uses the depth map to render the people poses in 3D. An overview of Pixels2Pose is displayed in Fig. \ref{fig:postprocessing}.  
It consists of three steps: first, a neural network called Pixels2Depth; second, a neural network called Depth2Pose; and finally, a post-processing module that combines the information from each network. Pixels2Depth processes the histogram coming from the sensor using 3D convolutional layers to render  depth maps with a resolution of 32$\times$32 pixels. Depth2Pose then processes this higher resolution depth map using 2D convolutional layers to output the 2D position of joints and limbs of all people present. This stage uses an adaptation of OpenPose \cite{OpenPose} specifically written for depth images rather than intensity images.  Finally, we associate the limb locations provided by Depth2Pose with the corresponding depth locations obtained from Pixels2Depth to recover distinct 3D skeletons of people. Further details on the different steps are provided in Supplementary Information 2.

\subsection*{Supervised training}
The two networks that we use for Pixels2Pose, Pixels2Depth and Depth2Pose, are each trained separately and then combined later. To train Pixels2Depth, we simultaneously record histograms from the vl53l5 sensor and the corresponding depth images with a Microsoft Azure Kinect DK.  The high-resolution images from the Kinect are downsampled to 32x32 pixels using bicubic interpolation before training.  We now have the data from the vl53l5 sensor and the corresponding ground truth depth label that can be used for training.  

To train Depth2Pose, we exploit the corresponding Kinect's RGB image, which is recorded at the same time as the depth image.  We can use the intensity images to extract 2D pose labels (confidence maps of joints and limbs position) via the RGB-based model OpenPose \cite{OpenPose}.  These 2D pose labels are the ground truth data used to train the Depth2Pose network.  During training, Depth2Pose learns the parameters of the network to convert a depth image from Pixels2Depth to the 2D pose obtained from the RGB image. After the supervised training of the networks, Pixels2Pose relies only on the vl53l5 sensor data with no additional camera necessary.

We trained three separate networks for reconstructing one, two, and three people in 3D.  We collected 7000 images for the training for the one-person network, 9500 for the two-person network, and 9500 for the three-person network. All the training and validation images are captured in a controlled laboratory environment. Further details on the network structure and on the training are provided in Supplementary Information 2. 

\subsection*{Pose estimation of multiple people in 3D}

\begin{figure*}[t!]
    \centering
    \begin{subfigure}[b]{0.20\textwidth}
    \caption{RGB Reference}
        \includegraphics[width=\textwidth]{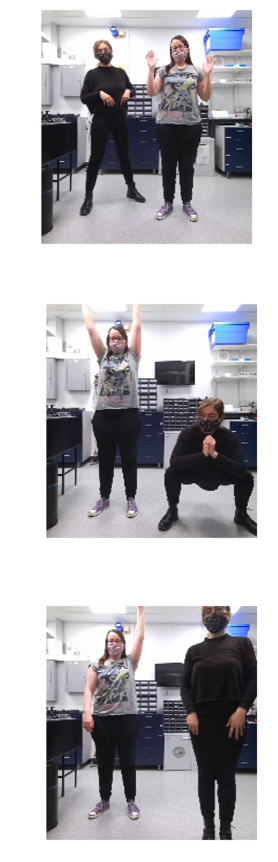}
       
    \end{subfigure}
~
    \begin{subfigure}[b]{0.20\textwidth}
     \caption{ Apparent resolution of the sensor}
        \includegraphics[width=\textwidth]{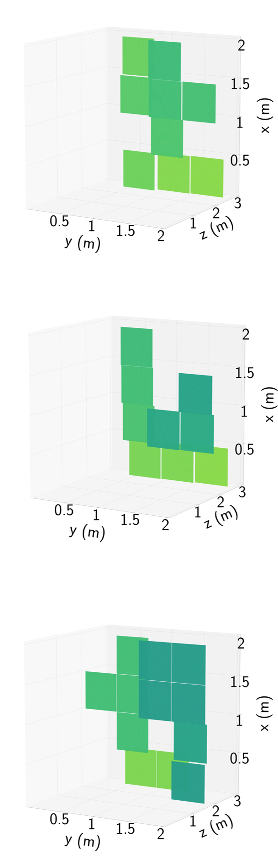}
        
    \end{subfigure}
~   
    \begin{subfigure}[b]{0.20\textwidth}
     \caption{ Pixels2Depth output}
        \includegraphics[width=\textwidth]{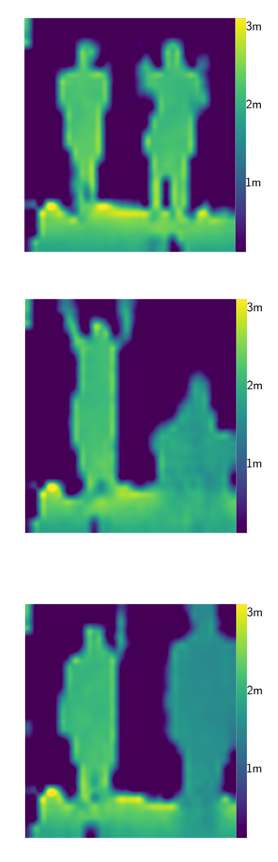}
        
    \end{subfigure}
~
     \begin{subfigure}[b]{0.20\textwidth}
     \caption{ Pixels2Pose output}
        \includegraphics[width=\textwidth]{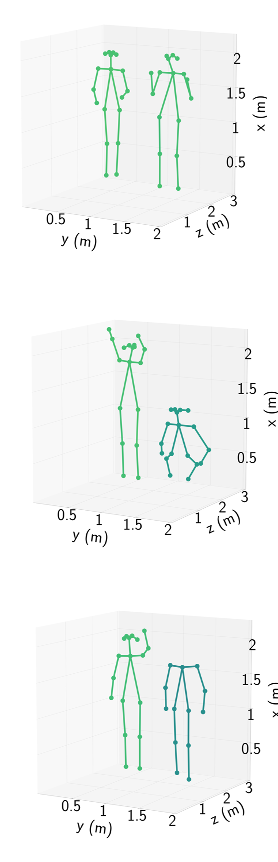}
       
    \end{subfigure}

    \caption{\textbf{Results with two people.} (a) is the RGB image taken by a Kinect for reference. (b) shows the 4 $\times$ 4 depth map corresponding to the maximum return of photon counts of the 4 $\times$ 4 $\times$ 100 histogram. (c) shows the output of Pixels2Depth. (d) shows the reconstruction of Pixels2Pose.}\label{fig:2people_res}
\end{figure*}

\begin{figure*}
\centering
\includegraphics[width=\textwidth]{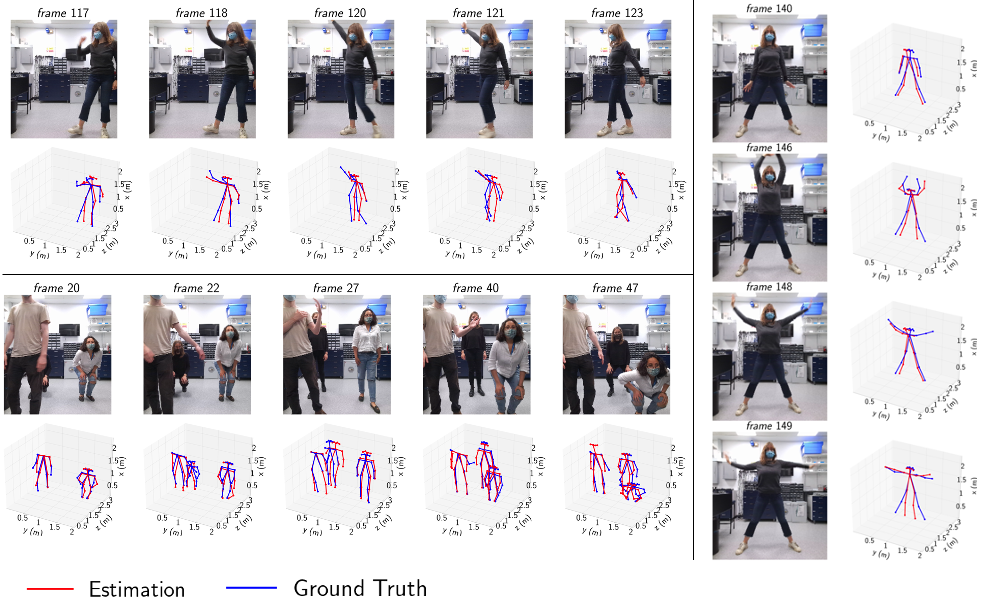}
\caption{\label{fig:Results_multi}
\textbf{Results with one and three people.} The 3D reconstructions on validation data and the corresponding RGB images are shown for different scenes containing one or three people.  }
\end{figure*}

Several frames showing the outputs of the two-person Pixels2Depth and Pixels2Pose networks are shown in Fig. \ref{fig:2people_res}.   We include the RGB image obtained from the Kinect as a reference of the input scene.  Note that this image was not used in any of the networks and is just shown as a guide for the reader.  Figure \ref{fig:Results_multi} shows several frames from the results for the one-person and three-person Pixels2Pose networks.  Here we also show the ground truth 3D pose as a reference for comparison.  

The ground truth 3D poses are obtained directly from the intensity and depth images of the Kinect.  We use the high-resolution RGB images and OpenPose \cite{OpenPose} to calculate the ground truth 2D pose data.  Each of the points in the 2D pose dataset corresponds to the $x, y$ location of a joint.  The $z$ location of each of the data points is obtained by using the corresponding depth information obtained from the associated Kinect depth image.

Supplementary information movies 1, 2, and 3 show videos of data obtained from the Pixels2Depth and Pixels2Pose networks for one, two, and three people, respectively. We also show the input to the network, the vl53l5 sensor data, and the reference data obtained from the Kinect camera.

\subsection*{Evaluation of performance}
We evaluate the accuracy of the estimated 3D poses on a validation dataset of 1500 images.  We use 500 frames for each scenario of one, two, and three people.  In Table \ref{table:errors}, we show the error in positions along the $x$, $y$, and $z$ axes for each joint in every pose that we estimate. The error is defined as the root mean squared error, expressed as (for the $x$-axis):
\begin{equation}
RMSE_{x} = \sqrt{\frac{1}{N}\sum_{i=1}^N (\widehat{x}_i - x_i)^2},
\end{equation}
with $N$ the number of validation frames, $(\widehat{x},\widehat{y},\widehat{z})$ the estimated positions, and $(x,y,z)$ the ground truth positions. We report the average error $AE$, defined as: 

\begin{equation}
AE = \frac{1}{N}\sum_{i=1}^N \sqrt{ (\widehat{x}_i - x_i)^2 + (\widehat{y}_i - y_i)^2 + (\widehat{z}_i - z_i)^2}. 
\end{equation}

We also report the percentages of correct key points (PCK-15, PCK-20, PCK-30), i.e.~the ratio of estimated body parts for which the distance to the ground truth is below 15, 20, and 30 cm respectively. We see that for the large core body parts i.e.~neck, shoulders, hips, and knees, more than 70\% of the estimates are within 15 cm of the real position; for the smaller body parts at the extremities, i.e.~ankle, wrists, and elbows, between 65\% and 90\% of estimates are within 30 cm.

\begin{table*}
\centering
\begin{tabular}{lrrrrrrr}
\hline
      &   $RMSE_x$ (cm)&   $RMSE_y$ (cm) &   $RMSE_z$ (cm)&  AE(cm) &  PCR-15 (\%) &   PCR-20 (\%) &   PCR-30 (\%)\\

\hline
 neck      &   5.4 &   6.0 &   8.5 &   9.5 &   80.0 &   88.0 &   92.0 \\
 shoulders &   5.8 &  12.4 &   9.2 &  12.3 &   72.5 &   80.2 &   86.3 \\
 hips      &   4.4 &   8.8 &   9.1 &  10.2 &   77.8 &   83.3 &   91.6 \\
 knees     &   5.6 &  11.1 &  10.1 &  11.9 &   72.1 &   81.7 &   89.8 \\
 ankles    &   7.9 &  15.1 &  11.3 &  15.1 &   62.1 &   74.4 &   86.4 \\
 elbows    &  17.7 &  19.9 &  13.4 &  19.6 &   60.9 &   68.6 &   75.4 \\
 wrists    &  22.6 &  26   &  17.6 &  25.9 &   50   &   57.5 &   65.1 \\
\hline
\end{tabular}
\caption{\textbf{Evaluation of the performance.} We report the root mean squared error between the estimated and the ground truth position of each joint for each axis x,y, and z. We also report the percentages of correct key points (PCK-15, PCK-20, PCK-30), i.e.~the ratio of estimated body parts for which the distance to the ground truth is below 15, 20, and 30 cm respectively.}
\label{table:errors}
\end{table*}

\begin{figure*}[t!]
    \centering
    \begin{subfigure}[b]{0.63\textwidth}
    \caption{Wrong arm position and no consistency in time}
        \includegraphics[width=\textwidth]{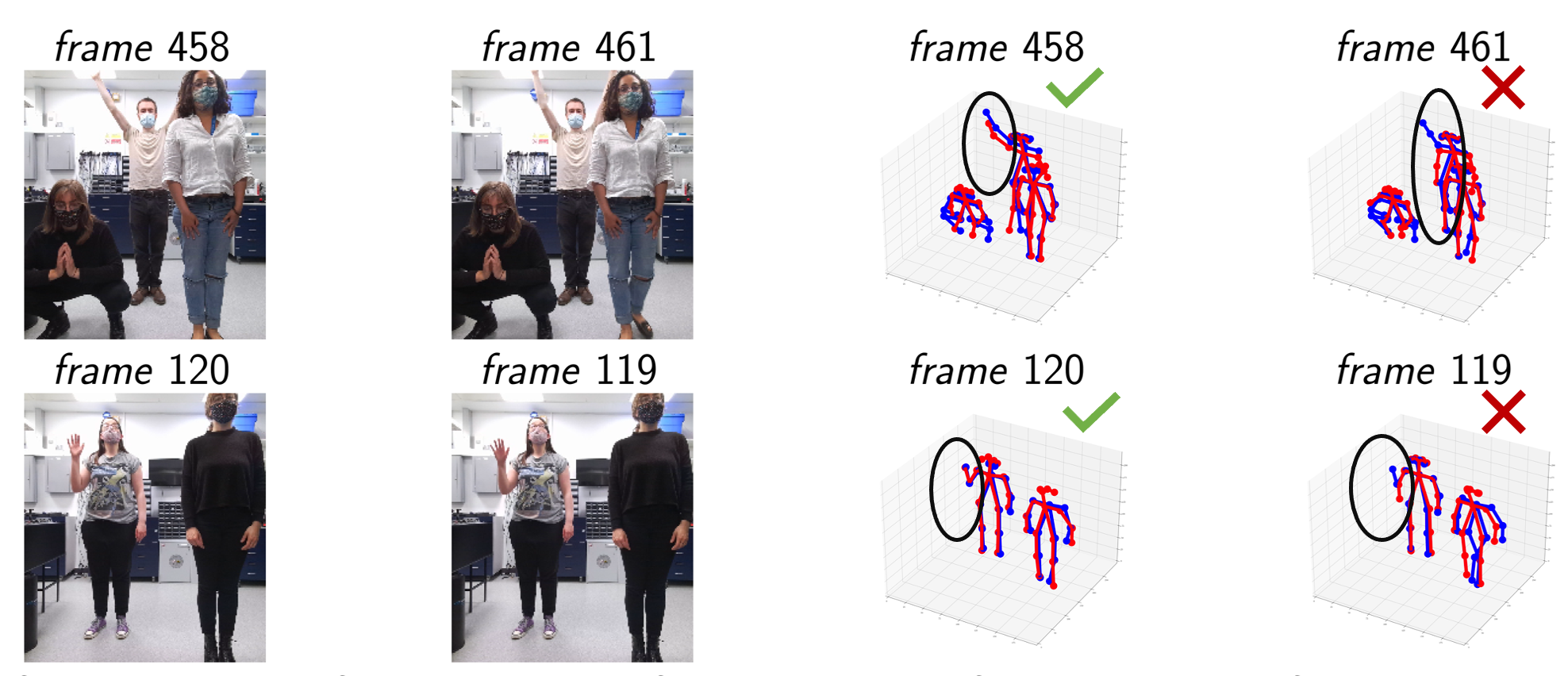}
    \end{subfigure}
~
    \begin{subfigure}[b]{0.31\textwidth}
     \caption{ Arms towards the sensor}
        \includegraphics[width=\textwidth]{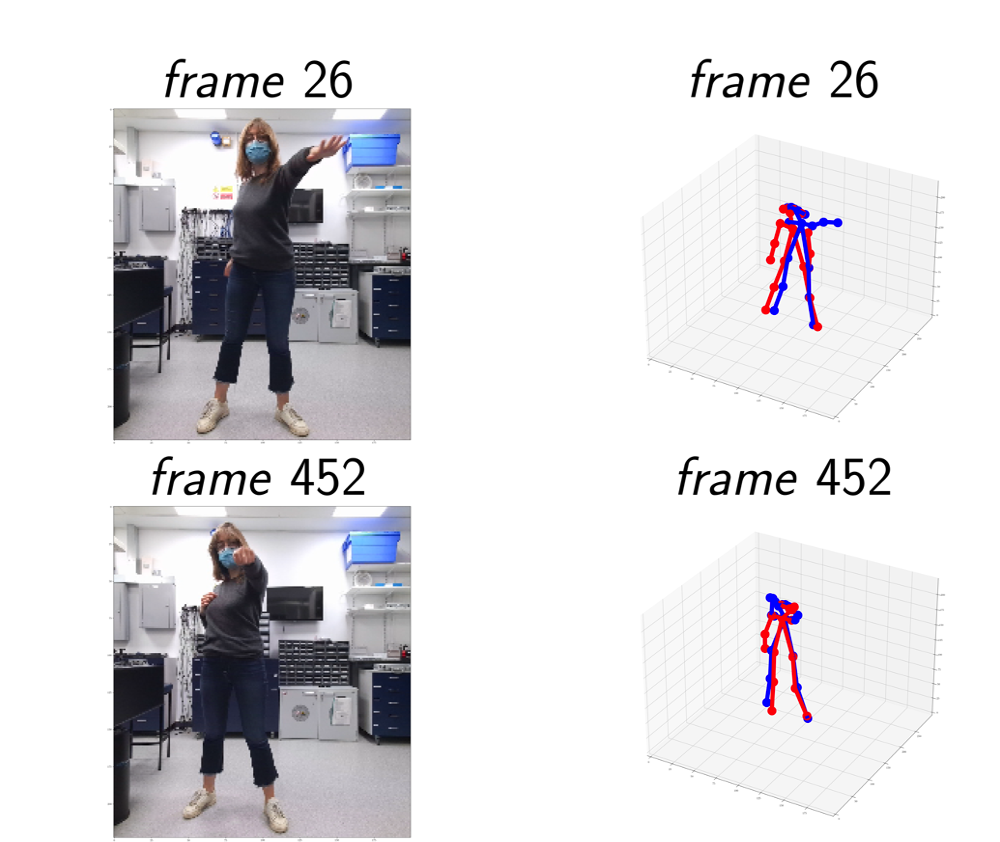}
    \end{subfigure}
~   
    \begin{subfigure}[b]{\textwidth}
     \caption{ Crossing}
        \includegraphics[width=\textwidth]{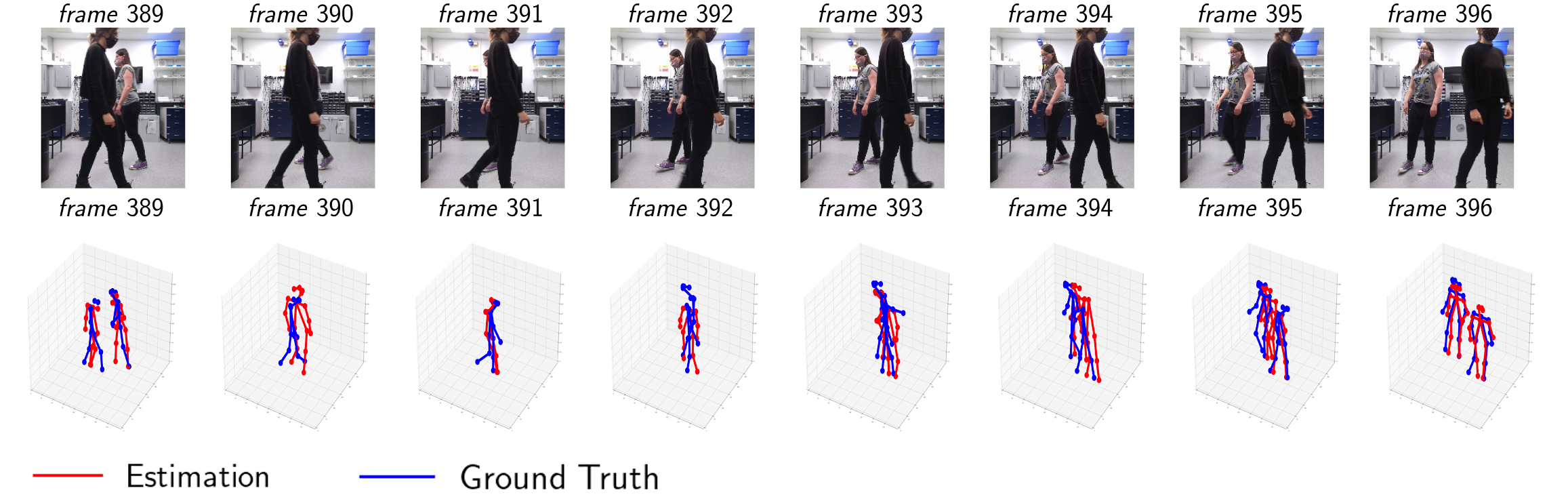}
    \end{subfigure}

    \caption{\textbf{Examples of failure cases.} (a) represents the case wrong arm position. (b) shows cases when arms were positioned in the axis of the sensor. (c) shows the issue when people are crossing. }\label{fig:failure_cases}
\end{figure*}

Supplementary movies 4, 5, and 6 show the reconstruction of poses from Pixels2Pose along with the ground truth obtained from the Kinect sensor. We see that the overall movement of the people is accurately recovered.

Fig. \ref{fig:failure_cases} shows examples of the most common failure cases of Pixels2Pose. The network could fail to identify arm movements when multiple people are present in the scene, e.g.~in the case of three people present, arms can be misplaced alongside the body, as in Fig. \ref{fig:failure_cases} (a). Moreover, movements over multiple time frames are sometimes unrealistic, e.g.~changes in the position of arms and legs that are too rapid are occasionally observed, as in Fig. \ref{fig:failure_cases} (a).  We also observe that people can disappear from the frame when crossing behind one another, as in Fig. \ref{fig:failure_cases} (c). The system might fail  to identify arms that are directed towards the sensor as in Fig. \ref{fig:failure_cases} (b).

\subsection*{Performance in other environments}

To demonstrate that the trained Pixels2Pose network is transferable between different environments, we took new data with the vl53l5 sensor in a new room and from two different angles.  No data from the second room was used in the training of the  Pixels2Pose network.  The acquired data was processed and 3D poses were reconstructed.  The results of this can be seen in the supplementary information 5. A video of the reconstruction in new environments is shown in the supplementary movie 7.  As with the training data captured from the vl53l5, the number of bins from the histogram was reduced from 144 to 100.  This ensures that there are no artefacts in the background that would affect the final result.

The data shows that the Pixels2Pose network recovers the 3D pose in an environment in which it was not trained, thus demonstrating the versatility of our system.  We note that in this case the average error of the body locations increases, and this is likely due to changes in the ambient light levels and the precise orientation and location of the vl53l5 sensor with respect to the subject.  These differences could be accounted for with further training of the network or a pre-processing step that corrects for orientation.

\subsection*{Computational requirements}

The model Pixels2Depth consists of 368,929 parameters of type float32 and requires about 4.7 MB of memory. The model Depth2Pose consists of 2,517,768 parameters of type float32 and 30 MB. For one frame, the processing time is 0.032s for Pixels2Depth, 0.032s for Depth2Pose, and 0.07s for the post-processing module, i.e.~the total processing time of Pixels2Pose is around 7 to 8 fps, processed on an NVidia Tesla RTX 6000 GPU. 

We can reduce the memory requirements of the networks using the Tensorflow Lite converter.  This can be used to create an appropriately sized network for implementation on computing systems with less resource than a GPU, e.g. mobile and IoT devices. Tensorflow Lite applies a post-training quantization to the trainable weights from floating-point to integer. After the conversion, the entire Pixels2Pose system requires only 5 MB of memory.  We find that the reduced-size networks have a very similar performance as the original models, often performing to within a few percent of the main network. The exact details of the performance of the Lite version of Pixels2Pose can be found in Supplementary Information 3. The lite models can be used directly on a Raspberry Pi 4, in real time together with the acquisition of the data. In this case, we can achieve a frame rate of 1 fps for both the acquisition and the processing of the data.

\section*{Discussion}
In this project, we have developed a machine learning approach to estimate poses of people in 3D from a cost-effective and compact time-of-flight sensor, containing only 4$\times$4 pixels. The sensor is small, light-weight, has a low power consumption, and can be easily integrated into consumer electronics such as smartphones or computers. The combined sensor and algorithm is capable of estimating the 3D poses of multiple humans in real-time at a maximum range of 3 m and at a frame rate of $\approx 7$ to 8 fps. 

This work shows the capabilities of low-cost ToF sensors to provide rich data from which key information can be extracted. This technology can be used for action/gesture recognition and will have applications in driver monitoring systems, human-computer interaction, and healthcare observation. We have detected large-scale objects in this work, and future work will focus on resolving finer features that will open up further applications, e.g.~facial structure for face ID applications, or finger and hand gestures for sign language identification.  The system could also be used for the reconstruction of more general shapes for simultaneous localization and mapping (SLAM), a navigation technique used by robots and autonomous vehicles. Furthermore, our 3D pose estimation system could be extended to other SPAD or RADAR detectors, including those used for non-line-of-sight (NLOS) imaging. 

We note that Pixel2Pose accurately tracks multiple humans in a 3D space, but it does not yet identify specific individuals within a scene.  That is to say, Pixels2Pose can track three people simultaneously, but it cannot label each of them separately.  This has obvious implications where data protection is an issue. It is not clear yet whether the current sensor would have the resolution in time and space to achieve accurate person identification, however, we note that neural networks have already been used to perform this task on people hidden from view \cite{Caramazza2018}.   This research direction will be of significant interest in the near future.

\section*{Method}

Our initial experimental setup for acquiring the training datasets consists of the vl53l5 sensor, mounted on a Raspberry Pi 3B, and a Microsoft Azure Kinect DK camera that records the reference RGB image and the reference depth image. The two sensors, the vl53l5 and Kinect, are placed as close as possible to each other to limit any paralax issues.  The radial lens distortion present in the Kinect depth image is corrected for.  This ensures that there is a one-to-one correspondence between the spatial locations of the pixels in the depth image and the RGB image. A picture of the setup used for training is shown in the supplementary information. 

As the Kinect sensor has a larger field of view than the vl53l5 sensor, we crop the Kinect depth and RGB images appropriately.  This means that the data provided to the network for training from the Kinect and vl53l5 sensor have the same field of view.  Both the Kinect and vl53l5 sensor operate at about 20 fps, however, the data for both is acquired asynchronously. To match the frames of both devices in time, we save the time at which each frame is recorded and post-process the data to have as close a match as possible. 

Up to three people walk in front of the sensors in random directions, in different positions, and with different arm gestures.
We recorded three different datasets containing one, two, or three persons. The one-person dataset contains 7500 frames, the two- and three-people datasets contain 11 000 frames each. In each case, the first 500 consecutive frames were set aside for validation. 
A picture of the setup can be found in Supplementary Information 1.

\bibliography   {references}

\begin{thebibliography}{10}
\urlstyle{rm}
\expandafter\ifx\csname url\endcsname\relax
  \def\url#1{\texttt{#1}}\fi
\expandafter\ifx\csname urlprefix\endcsname\relax\def\urlprefix{URL }\fi
\expandafter\ifx\csname doiprefix\endcsname\relax\def\doiprefix{DOI: }\fi
\providecommand{\bibinfo}[2]{#2}
\providecommand{\eprint}[2][]{\url{#2}}

\bibitem{:Zanfir}
\bibinfo{author}{Zanfir, M.}, \bibinfo{author}{Leordeanu, M.} \&
  \bibinfo{author}{Sminchisescu, C.}
\newblock \bibinfo{title}{The moving pose: An efficient 3d kinematics
  descriptor for low-latency action recognition and detection}.
\newblock In \emph{\bibinfo{booktitle}{2013 IEEE International Conference on
  Computer Vision}} (\bibinfo{year}{2013}).

\bibitem{Farooq:18}
\bibinfo{author}{Farooq, A.}, \bibinfo{author}{Jalal, A.} \&
  \bibinfo{author}{Kamal, S.}
\newblock \bibinfo{journal}{\bibinfo{title}{Dense rgb-d map-based human
  tracking and activity recognition using skin joints features and
  self-organizing map}}.
\newblock {\emph{\JournalTitle{KSII Transactions on Internet and Information
  Systems}}} \textbf{\bibinfo{volume}{9}} (\bibinfo{year}{2018}).

\bibitem{Dama:16}
\bibinfo{author}{Cippitelli, E.}, \bibinfo{author}{Gasparrini, S.},
  \bibinfo{author}{Gambi, E.} \& \bibinfo{author}{Spinsante, S.}
\newblock \bibinfo{journal}{\bibinfo{title}{A human activity recognition system
  using skeleton data from rgbd sensors}}.
\newblock {\emph{\JournalTitle{Computational Intelligence and Neuroscience}}}
  (\bibinfo{year}{2016}).

\bibitem{Mathis:18}
\bibinfo{author}{Mathis, A.} \emph{et~al.}
\newblock \bibinfo{journal}{\bibinfo{title}{{DeepLabCut: markerless pose
  estimation of user-defined body parts with deep learning}}}.
\newblock {\emph{\JournalTitle{{Nature Neuroscience}}}}
  \textbf{\bibinfo{volume}{{21}}} (\bibinfo{year}{{2018}}).

\bibitem{Moeslund:06}
\bibinfo{author}{Moeslund, T.~B.}, \bibinfo{author}{Hilton, A.} \&
  \bibinfo{author}{Kruger, V.}
\newblock \bibinfo{journal}{\bibinfo{title}{{A survey of advances in
  vision-based human motion capture and analysis}}}.
\newblock {\emph{\JournalTitle{{Computer Vision and Image Understanding}}}}
  (\bibinfo{year}{{2006}}).

\bibitem{Xiong_20}
\bibinfo{author}{Xiong, X.} \emph{et~al.}
\newblock \bibinfo{journal}{\bibinfo{title}{{S3D-CNN: skeleton-based 3D
  consecutive-low-pooling neural network for fall detection}}}.
\newblock {\emph{\JournalTitle{{Applied Intelligence}}}}
  \textbf{\bibinfo{volume}{{50}}} (\bibinfo{year}{{2020}}).

\bibitem{Bian_15}
\bibinfo{author}{Bian, Z.-P.}, \bibinfo{author}{Hou, J.},
  \bibinfo{author}{Chau, L.-P.} \& \bibinfo{author}{Magnenat-Thalmann, N.}
\newblock \bibinfo{journal}{\bibinfo{title}{{Fall Detection Based on Body Part
  Tracking Using a Depth Camera}}}.
\newblock {\emph{\JournalTitle{{IEEE Journal of Biomedical and Health
  Informatics}}}} \textbf{\bibinfo{volume}{{19}}} (\bibinfo{year}{{2015}}).

\bibitem{Serpa_20}
\bibinfo{author}{Serpa, Y.~R.}, \bibinfo{author}{Nogueira, M.~B.},
  \bibinfo{author}{Neto, P. P.~M.} \& \bibinfo{author}{Rodrigues, M. A.~F.}
\newblock \bibinfo{title}{Evaluating pose estimation as a solution to the fall
  detection problem}.
\newblock In \emph{\bibinfo{booktitle}{2020 IEEE International Conference on
  Serious Games and Applications for Health}} (\bibinfo{year}{2020}).

\bibitem{Qingqiang_21}
\bibinfo{author}{Wu, Q.}, \bibinfo{author}{Xu, G.}, \bibinfo{author}{Wei, F.},
  \bibinfo{author}{Chen, L.} \& \bibinfo{author}{Zhang, S.}
\newblock \bibinfo{journal}{\bibinfo{title}{Rgb-d videos-based early prediction
  of infant cerebral palsy via general movements complexity}}.
\newblock {\emph{\JournalTitle{IEEE Access}}} \textbf{\bibinfo{volume}{9}}
  (\bibinfo{year}{2021}).

\bibitem{Physical_therapy_19}
\bibinfo{author}{Gu, Y.} \emph{et~al.}
\newblock \bibinfo{title}{Home-based physical therapy with an interactive
  computer vision system}.
\newblock In \emph{\bibinfo{booktitle}{2019 IEEE/CVF International Conference
  on Computer Vision Workshop}} (\bibinfo{year}{2019}).

\bibitem{fallrecoveryassist16}
\bibinfo{author}{Withanage, K.~I.}, \bibinfo{author}{Lee, I.},
  \bibinfo{author}{Brinkworth, R.}, \bibinfo{author}{Mackintosh, S.} \&
  \bibinfo{author}{Thewlis, D.}
\newblock \bibinfo{journal}{\bibinfo{title}{Fall recovery subactivity
  recognition with rgb-d cameras}}.
\newblock {\emph{\JournalTitle{IEEE Transactions on Industrial Informatics}}}
  \textbf{\bibinfo{volume}{12}} (\bibinfo{year}{2016}).

\bibitem{monitoringsleep_18}
\bibinfo{author}{Torres, C.}, \bibinfo{author}{Fried, J.~C.},
  \bibinfo{author}{Rose, K.} \& \bibinfo{author}{Manjunath, B.~S.}
\newblock \bibinfo{journal}{\bibinfo{title}{A multiview multimodal system for
  monitoring patient sleep}}.
\newblock {\emph{\JournalTitle{IEEE Transactions on Multimedia}}}
  \textbf{\bibinfo{volume}{20}} (\bibinfo{year}{2018}).

\bibitem{Park_17}
\bibinfo{author}{Park, S.}, \bibinfo{author}{Chang, J.~Y.},
  \bibinfo{author}{Jeong, H.}, \bibinfo{author}{Lee, J.-H.} \&
  \bibinfo{author}{Park, J.-Y.}
\newblock \bibinfo{title}{{Accurate and Efficient 3D Human Pose Estimation
  Algorithm using Single Depth Images for Pose Analysis in Golf}}.
\newblock In \emph{\bibinfo{booktitle}{{2017 IEEE Conference on Computer Vision
  and Pattern Recognition Workshops}}} (\bibinfo{year}{{2017}}).

\bibitem{Lewandowski_19}
\bibinfo{author}{Lewandowski, B.}, \bibinfo{author}{Liebner, J.},
  \bibinfo{author}{Wengefeld, T.}, \bibinfo{author}{Müller, S.} \&
  \bibinfo{author}{Gross, H.-M.}
\newblock \bibinfo{title}{Fast and robust 3d person detector and posture
  estimator for mobile robotic applications}.
\newblock In \emph{\bibinfo{booktitle}{2019 International Conference on
  Robotics and Automation}} (\bibinfo{year}{2019}).

\bibitem{Yang_19}
\bibinfo{author}{Yang, Z.}, \bibinfo{author}{Li, Y.}, \bibinfo{author}{Yang,
  J.} \& \bibinfo{author}{Luo, J.}
\newblock \bibinfo{journal}{\bibinfo{title}{Action recognition with
  spatio–temporal visual attention on skeleton image sequences}}.
\newblock {\emph{\JournalTitle{IEEE Transactions on Circuits and Systems for
  Video Technology}}} \textbf{\bibinfo{volume}{29}} (\bibinfo{year}{2019}).

\bibitem{Keçeli_17}
\bibinfo{author}{Keçeli, A.~S.}, \bibinfo{author}{Kaya, A.} \&
  \bibinfo{author}{Can, A.~B.}
\newblock \bibinfo{title}{Action recognition with skeletal volume and deep
  learning}.
\newblock In \emph{\bibinfo{booktitle}{2017 25th Signal Processing and
  Communications Applications Conference}} (\bibinfo{year}{2017}).

\bibitem{Liu:2017}
\bibinfo{author}{Liu, J.}, \bibinfo{author}{Rahmani, H.},
  \bibinfo{author}{Akhtar, N.} \& \bibinfo{author}{Mian, A.}
\newblock \bibinfo{journal}{\bibinfo{title}{Learning human pose models from
  synthesized data for robust rgb-d action recognition}}.
\newblock {\emph{\JournalTitle{International Journal of Computer Vision}}}
  \textbf{\bibinfo{volume}{127}} (\bibinfo{year}{2019}).

\bibitem{Baldi-20}
\bibinfo{author}{Baldi, T.~L.}, \bibinfo{author}{Farina, F.},
  \bibinfo{author}{Garulli, A.}, \bibinfo{author}{Giannitrapani, A.} \&
  \bibinfo{author}{Prattichizzo, D.}
\newblock \bibinfo{journal}{\bibinfo{title}{{Upper Body Pose Estimation Using
  Wearable Inertial Sensors and Multiplicative Kalman Filter}}}.
\newblock {\emph{\JournalTitle{{IEEE Sensors Journal}}}}
  \textbf{\bibinfo{volume}{{20}}} (\bibinfo{year}{{2020}}).

\bibitem{Yun_06}
\bibinfo{author}{Yun, X.} \& \bibinfo{author}{Bachmann, E.~R.}
\newblock \bibinfo{journal}{\bibinfo{title}{{Design, implementation, and
  experimental results of a quaternion-based Kalman filter for human body
  motion tracking}}}.
\newblock {\emph{\JournalTitle{{IEEE Transactions on Robotics}}}}
  \textbf{\bibinfo{volume}{{22}}} (\bibinfo{year}{{2006}}).

\bibitem{Marcard_18}
\bibinfo{author}{von Marcard, T.}, \bibinfo{author}{Henschel, R.},
  \bibinfo{author}{Black, M.~J.}, \bibinfo{author}{Rosenhahn, B.} \&
  \bibinfo{author}{Pons-Moll, G.}
\newblock \bibinfo{title}{{Recovering Accurate 3D Human Pose in the Wild Using
  IMUs and a Moving Camera}}.
\newblock In \emph{\bibinfo{booktitle}{{Computer Vision - ECCV 2018, PT X}}},
  vol. \bibinfo{volume}{{11214}} (\bibinfo{year}{{2018}}).

\bibitem{Gilbert_19}
\bibinfo{author}{Gilbert, A.}, \bibinfo{author}{Trumble, M.},
  \bibinfo{author}{Malleson, C.}, \bibinfo{author}{Hilton, A.} \&
  \bibinfo{author}{Collomosse, J.}
\newblock \bibinfo{journal}{\bibinfo{title}{{Fusing Visual and Inertial Sensors
  with Semantics for 3D Human Pose Estimation}}}.
\newblock {\emph{\JournalTitle{{International Journal of Computer Vision}}}}
  \textbf{\bibinfo{volume}{{127}}} (\bibinfo{year}{{2019}}).

\bibitem{Bodysensor_sports}
\bibinfo{author}{Aminian, K.} \& \bibinfo{author}{Najafi, B.}
\newblock \bibinfo{journal}{\bibinfo{title}{Capturing human motion using
  body-fixed sensors: Outdoor measurement and clinical applications}}.
\newblock {\emph{\JournalTitle{Journal of Visualization and Computer
  Animation}}}  (\bibinfo{year}{2004}).

\bibitem{Bodysensors4}
\bibinfo{author}{De-Magalhaes, F.~A.}, \bibinfo{author}{Vannozzi, G.},
  \bibinfo{author}{Gatta, G.} \& \bibinfo{author}{Fantozzi, S.}
\newblock \bibinfo{journal}{\bibinfo{title}{Wearable inertial sensors in
  swimming motion analysis: a systematic review}}.
\newblock {\emph{\JournalTitle{Journal of Sports Sciences}}}
  \textbf{\bibinfo{volume}{33}} (\bibinfo{year}{2014}).

\bibitem{BodySensors3}
\bibinfo{author}{Eckardt, F.}, \bibinfo{author}{Münz, A.} \&
  \bibinfo{author}{Witte, K.}
\newblock \bibinfo{journal}{\bibinfo{title}{Application of a full body inertial
  measurement system in dressage riding}}.
\newblock {\emph{\JournalTitle{Journal of Equine Veterinary Science}}}
  \textbf{\bibinfo{volume}{34}} (\bibinfo{year}{2014}).

\bibitem{Vlasic_08}
\bibinfo{author}{Vlasic, D.}, \bibinfo{author}{Baran, I.},
  \bibinfo{author}{Matusik, W.} \& \bibinfo{author}{Popovic, J.}
\newblock \bibinfo{journal}{\bibinfo{title}{{Articulated mesh animation from
  multi-view silhouettes}}}.
\newblock {\emph{\JournalTitle{{ACM Trans. Graph.}}}}
  \textbf{\bibinfo{volume}{{27}}} (\bibinfo{year}{{2008}}).

\bibitem{Carranza_03}
\bibinfo{author}{Carranza, J.}, \bibinfo{author}{Theobalt, C.},
  \bibinfo{author}{Magnor, M.} \& \bibinfo{author}{Seidel, H.}
\newblock \bibinfo{journal}{\bibinfo{title}{{Free-viewpoint video of human
  actors}}}.
\newblock {\emph{\JournalTitle{{ACM Trans. Graph.}}}}
  \textbf{\bibinfo{volume}{{22}}} (\bibinfo{year}{{2003}}).

\bibitem{Iskakov_19}
\bibinfo{author}{Iskakov, K.}, \bibinfo{author}{Burkov, E.},
  \bibinfo{author}{Lempitsky, V.} \& \bibinfo{author}{Malkov, Y.}
\newblock \bibinfo{title}{{Learnable Triangulation of Human Pose}}.
\newblock In \emph{\bibinfo{booktitle}{{2019 IEEE/CVF International Conference
  on Computer Vision}}} (\bibinfo{year}{{2019}}).

\bibitem{Mehrizi_18}
\bibinfo{author}{Mehrizi, R.} \emph{et~al.}
\newblock \bibinfo{title}{Toward marker-free {3D} pose estimation in lifting: A
  deep multi-view solution}.
\newblock In \emph{\bibinfo{booktitle}{Proceedings - 13th IEEE International
  Conference on Automatic Face and Gesture Recognition, FG 2018}}
  (\bibinfo{year}{2018}).

\bibitem{Body_markers_avatar}
\bibinfo{author}{Lee, J.}, \bibinfo{author}{Chai, J.},
  \bibinfo{author}{Reitsma, P. S.~A.}, \bibinfo{author}{Hodgins, J.~K.} \&
  \bibinfo{author}{Pollard, N.~S.}
\newblock \bibinfo{title}{Interactive control of avatars animated with human
  motion data}.
\newblock In \emph{\bibinfo{booktitle}{Proceedings of the 29th Annual
  Conference on Computer Graphics and Interactive Techniques}}
  (\bibinfo{year}{2002}).

\bibitem{Poppe_07}
\bibinfo{author}{Poppe, R.}
\newblock \bibinfo{journal}{\bibinfo{title}{{Vision-based human motion
  analysis: An overview}}}.
\newblock {\emph{\JournalTitle{{Computer Vision and Image Understanding}}}}
  \textbf{\bibinfo{volume}{{108}}} (\bibinfo{year}{{2007}}).

\bibitem{Moeslund_06}
\bibinfo{author}{Moeslund, T.~B.}, \bibinfo{author}{Hilton, A.} \&
  \bibinfo{author}{Kruger, V.}
\newblock \bibinfo{journal}{\bibinfo{title}{{A survey of advances in
  vision-based human motion capture and analysis}}}.
\newblock {\emph{\JournalTitle{{Computer Vision and Image Understanding}}}}
  \textbf{\bibinfo{volume}{{104}}} (\bibinfo{year}{{2006}}).

\bibitem{Bala_20}
\bibinfo{author}{Bala, P.~C.} \emph{et~al.}
\newblock \bibinfo{journal}{\bibinfo{title}{{Automated markerless pose
  estimation in freely moving macaques with OpenMonkeyStudio}}}.
\newblock {\emph{\JournalTitle{{Nature Communications}}}}
  \textbf{\bibinfo{volume}{{11}}} (\bibinfo{year}{{2020}}).

\bibitem{Kidzinski_20}
\bibinfo{author}{Kidzinski, L.} \emph{et~al.}
\newblock \bibinfo{journal}{\bibinfo{title}{{Deep neural networks enable
  quantitative movement analysis using single-camera videos}}}.
\newblock {\emph{\JournalTitle{{Nature Communications}}}}
  \textbf{\bibinfo{volume}{{11}}} (\bibinfo{year}{{2020}}).

\bibitem{Rogez_20}
\bibinfo{author}{Rogez, G.}, \bibinfo{author}{Weinzaepfel, P.} \&
  \bibinfo{author}{Schmid, C.}
\newblock \bibinfo{journal}{\bibinfo{title}{{LCR-Net plus plus : Multi-Person
  2D and 3D Pose Detection in Natural Images}}}.
\newblock {\emph{\JournalTitle{{IEEE Transactions on Pattern Analysis and
  Machine Intelligence}}}} \textbf{\bibinfo{volume}{{42}}}
  (\bibinfo{year}{{2020}}).

\bibitem{Mehta_20}
\bibinfo{author}{Mehta, D.} \emph{et~al.}
\newblock \bibinfo{journal}{\bibinfo{title}{{XNect: Real-time Multi-Person 3D
  Motion Capture with a Single RGB Camera}}}.
\newblock {\emph{\JournalTitle{{ACM Trans. Graph.}}}}
  \textbf{\bibinfo{volume}{{39}}} (\bibinfo{year}{{2020}}).

\bibitem{Benzine_21}
\bibinfo{author}{Benzine, A.}, \bibinfo{author}{Luvison, B.},
  \bibinfo{author}{Pham, Q.~C.} \& \bibinfo{author}{Achard, C.}
\newblock \bibinfo{journal}{\bibinfo{title}{{Single-shot 3D multi-person pose
  estimation in complex images}}}.
\newblock {\emph{\JournalTitle{{Pattern Recognition}}}}
  \textbf{\bibinfo{volume}{{112}}} (\bibinfo{year}{{2021}}).

\bibitem{Liu_20}
\bibinfo{author}{Liu, J.} \emph{et~al.}
\newblock \bibinfo{journal}{\bibinfo{title}{{Feature Boosting Network For 3D
  Pose Estimation}}}.
\newblock {\emph{\JournalTitle{{IEEE Transactions on Pattern Analysis and
  Machine Intelligence}}}} \textbf{\bibinfo{volume}{{42}}}
  (\bibinfo{year}{{2020}}).

\bibitem{Agarwal_06}
\bibinfo{author}{Agarwal, A.} \& \bibinfo{author}{Triggs, B.}
\newblock \bibinfo{journal}{\bibinfo{title}{Recovering 3d human pose from
  monocular images}}.
\newblock {\emph{\JournalTitle{IEEE Transactions on Pattern Analysis and
  Machine Intelligence}}} \textbf{\bibinfo{volume}{28}} (\bibinfo{year}{2006}).

\bibitem{Chen_17}
\bibinfo{author}{Chen, C.-H.} \& \bibinfo{author}{Ramanan, D.}
\newblock \bibinfo{title}{{3D Human Pose Estimation=2D Pose Estimation plus
  Matching}}.
\newblock In \emph{\bibinfo{booktitle}{{2017 IEEE Conference on Computer Vision
  and Pattern Recognition}}} (\bibinfo{year}{{2017}}).

\bibitem{Wang2020}
\bibinfo{author}{Zhe, W.}
\newblock
  \bibinfo{howpublished}{https://github.com/wangzheallen/awesome-human-pose-estimation}
  (\bibinfo{year}{2020}).

\bibitem{Zhou_20}
\bibinfo{author}{Zhou, Y.}, \bibinfo{author}{Dong, H.} \&
  \bibinfo{author}{Saddik, A.~E.}
\newblock \bibinfo{journal}{\bibinfo{title}{{Learning to Estimate 3D Human Pose
  From Point Cloud}}}.
\newblock {\emph{\JournalTitle{{IEEE Sensors Journal}}}}
  \textbf{\bibinfo{volume}{{20}}} (\bibinfo{year}{{2020}}).

\bibitem{Zhang_20}
\bibinfo{author}{Zhang, Z.}, \bibinfo{author}{Hu, L.}, \bibinfo{author}{Deng,
  X.} \& \bibinfo{author}{Xia, S.}
\newblock \bibinfo{journal}{\bibinfo{title}{{Weakly Supervised Adversarial
  Learning for 3D Human Pose Estimation from Point Clouds}}}.
\newblock {\emph{\JournalTitle{{IEEE Transactions on Visualization and Computer
  Graphics}}}} \textbf{\bibinfo{volume}{{26}}} (\bibinfo{year}{{2020}}).

\bibitem{Moon_18}
\bibinfo{author}{Moon, G.}, \bibinfo{author}{Chang, J.~Y.} \&
  \bibinfo{author}{Lee, K.~M.}
\newblock \bibinfo{title}{{V2V-PoseNet: Voxel-to-Voxel Prediction Network for
  Accurate 3D Hand and Human Pose Estimation from a Single Depth Map}}.
\newblock In \emph{\bibinfo{booktitle}{{2018 IEEE/CVF Conference on Computer
  Vision and Pattern Recognition}}} (\bibinfo{year}{{2018}}).

\bibitem{survey_depthimagery_13}
\bibinfo{author}{Chen, L.}, \bibinfo{author}{Wei, H.} \&
  \bibinfo{author}{Ferryman, J.}
\newblock \bibinfo{journal}{\bibinfo{title}{A survey of human motion analysis
  using depth imagery}}.
\newblock {\emph{\JournalTitle{Pattern Recognition Letters}}}
  \textbf{\bibinfo{volume}{34}} (\bibinfo{year}{2013}).

\bibitem{Sun_16}
\bibinfo{author}{Sun, M.-J.} \emph{et~al.}
\newblock \bibinfo{journal}{\bibinfo{title}{{Single-pixel three-dimensional
  imaging with time-based depth resolution}}}.
\newblock {\emph{\JournalTitle{{Nature Communications}}}}
  \textbf{\bibinfo{volume}{{7}}} (\bibinfo{year}{{2016}}).

\bibitem{Zhang_15}
\bibinfo{author}{Zhang, Z.}, \bibinfo{author}{Ma, X.} \&
  \bibinfo{author}{Zhong, J.}
\newblock \bibinfo{journal}{\bibinfo{title}{{Single-pixel imaging by means of
  Fourier spectrum acquisition}}}.
\newblock {\emph{\JournalTitle{{Nature Communications}}}}
  \textbf{\bibinfo{volume}{{6}}} (\bibinfo{year}{{2015}}).

\bibitem{Turpin:20}
\bibinfo{author}{Turpin, A.} \emph{et~al.}
\newblock \bibinfo{journal}{\bibinfo{title}{Spatial images from temporal
  data}}.
\newblock {\emph{\JournalTitle{Optica}}} \textbf{\bibinfo{volume}{7}}
  (\bibinfo{year}{2020}).

\bibitem{Turpin_21}
\bibinfo{author}{Turpin, A.} \emph{et~al.}
\newblock \bibinfo{journal}{\bibinfo{title}{3d imaging from multipath temporal
  echoes}}.
\newblock {\emph{\JournalTitle{Phys. Rev. Lett.}}}
  \textbf{\bibinfo{volume}{126}} (\bibinfo{year}{2021}).

\bibitem{Zhao_18}
\bibinfo{author}{Zhao, M.} \emph{et~al.}
\newblock \bibinfo{title}{Rf-based 3d skeletons}.
\newblock In \emph{\bibinfo{booktitle}{Proceedings of the 2018 Conference of
  the ACM Special Interest Group on Data Communication}}
  (\bibinfo{year}{2018}).

\bibitem{Zhao_18_2D}
\bibinfo{author}{Zhao, M.} \emph{et~al.}
\newblock \bibinfo{title}{Through-wall human pose estimation using radio
  signals}.
\newblock In \emph{\bibinfo{booktitle}{2018 IEEE/CVF Conference on Computer
  Vision and Pattern Recognition (CVPR)}} (\bibinfo{year}{2018}).

\bibitem{Wetzstein_20}
\bibinfo{author}{Nishimura, M.}, \bibinfo{author}{Lindell, D.~B.},
  \bibinfo{author}{Metzler, C.} \& \bibinfo{author}{Wetzstein, G.}
\newblock \bibinfo{title}{Disambiguating monocular depth estimation with a
  single transient}.
\newblock In \emph{\bibinfo{booktitle}{Computer Vision -- ECCV 2020}}
  (\bibinfo{publisher}{Springer International Publishing},
  \bibinfo{year}{2020}).

\bibitem{Lindell:2018}
\bibinfo{author}{Lindell, D.~B.}, \bibinfo{author}{O’Toole, M.} \&
  \bibinfo{author}{Wetzstein, G.}
\newblock \bibinfo{journal}{\bibinfo{title}{{Single-Photon 3D Imaging with Deep
  Sensor Fusion}}}.
\newblock {\emph{\JournalTitle{ACM Trans. Graph.}}}  (\bibinfo{year}{2018}).

\bibitem{Sun_20}
\bibinfo{author}{Sun, Q.} \emph{et~al.}
\newblock \bibinfo{journal}{\bibinfo{title}{End-to-end learned, optically coded
  super-resolution spad camera}}.
\newblock {\emph{\JournalTitle{ACM Trans. Graph.}}}
  \textbf{\bibinfo{volume}{39}} (\bibinfo{year}{2020}).

\bibitem{Ruget:21}
\bibinfo{author}{Ruget, A.} \emph{et~al.}
\newblock \bibinfo{journal}{\bibinfo{title}{Robust super-resolution depth
  imaging via a multi-feature fusion deep network}}.
\newblock {\emph{\JournalTitle{Opt. Express}}} \textbf{\bibinfo{volume}{29}}
  (\bibinfo{year}{2021}).

\bibitem{Callenberg2021CheapSPAD}
\bibinfo{author}{Callenberg, C.}, \bibinfo{author}{Shi, Z.},
  \bibinfo{author}{Heide, F.} \& \bibinfo{author}{Hullin, M.~B.}
\newblock \bibinfo{journal}{\bibinfo{title}{Low-cost spad sensing for
  non-line-of-sight tracking, material classification and depth imaging}}.
\newblock {\emph{\JournalTitle{ACM Trans. Graph.}}}
  \textbf{\bibinfo{volume}{40}} (\bibinfo{year}{2021}).

\bibitem{OpenPose}
\bibinfo{author}{Cao, Z.}, \bibinfo{author}{Simon, T.}, \bibinfo{author}{Wei,
  S.-E.} \& \bibinfo{author}{Sheikh, Y.}
\newblock \bibinfo{title}{Realtime multi-person 2d pose estimation using part
  affinity fields}.
\newblock In \emph{\bibinfo{booktitle}{2017 IEEE Conference on Computer Vision
  and Pattern Recognition}} (\bibinfo{year}{2017}).

\bibitem{Caramazza2018}
\bibinfo{author}{Caramazza, P.} \emph{et~al.}
\newblock \bibinfo{journal}{\bibinfo{title}{Neural network identification of
  people hidden from view with a single-pixel, single-photon detector}}.
\newblock {\emph{\JournalTitle{Scientific Reports}}}
  \textbf{\bibinfo{volume}{8}} (\bibinfo{year}{2018}).

\end{thebibliography}


\section*{Code availability}

Code for generating the 3D pose from the ToF sensor can be found at https://github.com/HWQuantum/Real-time-low-cost-multi-person-3D-pose-estimation.

\section*{Funding}

This work was supported by EPSRC through grants  EP/S001638/1 and EP/T00097X/1. Also it is supported by the UK Royal Academy of Engineering Research Fellowship Scheme (Project
RF/201718/17128) and DSTL Dasa project DSTLX1000147844).

\section*{Acknowledgements} 

We thank the authors of OpenPose \cite{OpenPose} for their code. We thank Frédéric Ruget for his guidance on the figures.

\section*{Author contributions statement}
A.R, B.H, A.H, and J.L conceived the experiment. A.R, M.T, G.M, B.H, and J.L conducted the experiment.  A.R implemented the algorithm. F.Z, S.S, I.G, S.M, A.H, and J.L contributed to the algorithm development. All authors contributed to the writing and reviewing of the manuscript.  

\section*{Competing interests} 

The authors declare no competing interests.

\end{document}